\documentclass[11pt,a4paper]{article}
\usepackage[hyperref]{acl2019}
\usepackage{times}
\usepackage{latexsym}

\usepackage{url}

\usepackage{graphicx}
\usepackage{amssymb}
\usepackage{amsmath}
\usepackage{multirow}

\aclfinalcopy 

\title{The Missing Ingredient in Zero-Shot Neural Machine Translation}

\author{Naveen Arivazhagan, Ankur Bapna, Orhan Firat, Roee Aharoni, Melvin Johnson, Wolfgang Macherey \\
\{navari,ankurbpn,orhanf,wmach\}@google.com\\
Google AI \\
}

\date{}

\begin{document}
\maketitle
\begin{abstract}
Multilingual Neural Machine Translation (NMT) models are capable of translating between multiple source and target languages. Despite various approaches to train such models, they have difficulty with zero-shot translation: translating between language pairs that were not together seen during training. In this paper we first diagnose why state-of-the-art multilingual NMT models that rely purely on parameter sharing, fail to generalize to unseen language pairs. We then propose auxiliary losses on the NMT encoder that impose representational invariance across languages. Our simple approach vastly improves zero-shot translation quality without regressing on supervised directions. For the first time, on WMT14 English-French-German, we achieve zero-shot performance that is on par with pivoting. We also demonstrate the easy scalability of our approach to multiple languages on the IWSLT 2017 shared task.
\end{abstract}

\section{Introduction}
    Neural Machine Translation (NMT) \cite{DBLP:journals/corr/SutskeverVL14,DBLP:journals/corr/BahdanauCB14,DBLP:journals/corr/ChoMGBSB14} allows for a simple extension to the multilingual setting with the ultimate goal of a single model supporting translation between all languages \cite{dong2015multi,luong2015multi,DBLP:journals/corr/FiratCB16,johnson2016google}. The challenge, however, is that parallel training data is usually only available in concurrence with English. Even so, it seems plausible that, given good cross-language generalization, the model should be able to translate between any pairing of supported source and target languages - even untrained, non-English pairs. However, despite the model's excellent performance on supervised directions, the quality on these zero-shot directions consistently lags behind pivoting by 2-10 BLEU points \citep{DBLP:conf/emnlp/FiratSAYC16,johnson2016google,ha2017effective,lu2018neural}.

    The failure of multilingual models to generalize to these zero-shot directions has been patched up by using a few techniques. Typically, translation between zero-shot language pairs is instead achieved by following a two-step process of pivoting or bridging through a common language \cite{Wu:2007:PLA:1452924.1452934,salloum2013dialectal}. Decoding through two noisy channels, however, doubles the latency and compounds errors. Several data augmentation, or zero resource, methods have therefore been proposed to enable one-step translation \cite{DBLP:conf/emnlp/FiratSAYC16,DBLP:journals/corr/ChenLCL17}, but these require multiple training phases and grow quadratically in the number of languages. In this work, we try to understand the generalization problem that impairs zero-shot translation and resolve it directly rather than treating the symptoms.

    The success of zero-shot translation depends on the ability of the model to learn language invariant features, or an \textit{interlingua}, for cross-lingual transfer \citep{ben2007analysis,mansour2009domain}. We begin with an error analysis which reveals that the standard approach of tying weights in the encoder is, by itself, not a sufficient constraint to elicit this, and the model enters a failure mode when translating between zero-shot languages \cite{DBLP:conf/emnlp/FiratSAYC16}.

    To resolve this issue, we begin to view zero-shot translation as a domain adaptation problem \cite{ben2007analysis,mansour2009domain} in multilingual NMT. We treat English, the language with which we always have parallel data, as the source domain, and the other languages collectively as the target domain. Rather than passively relying on parameter sharing, we apply auxiliary losses to explicitly incentivize the model to use domain/source-language invariant representations. By essentially using English representations as an implicit pivot in the continuous latent space, we achieve large improvements on zero-shot translation performance.

    As we demonstrate on WMT14 (English-German-French), we are for the first time able to achieve zero-shot performance that is on par with pivoting. We are able to do this without any meaningful regression on the supervised directions and without the multi-phase training and quadratic complexity of data-synthesis approaches. We show how our approach can easily be scaled up to more languages on IWSLT17. Our results suggest that explicitly incentivizing cross-lingual transfer may be the missing ingredient to improving the quality of zero-shot translation.

\section{Related Work}
    \paragraph{Multilingual NMT} was first proposed by \citet{dong2015multi} for translating from one source language to multiple target languages. Subsequently, sequence-to-sequence models were extended to the many-to-many setting by the addition of task-specific encoder and decoder modules \cite{luong2015multi,DBLP:conf/emnlp/FiratSAYC16}. Since then, \citet{johnson2016google, ha2016toward} have shown that a vanilla sequence-to-sequence model with a single encoder and decoder can be used in the many-to-many setting by using a special token to indicate the target language.
    
    \paragraph{Zero-Shot Translation} was first demonstrated by \citet{johnson2016google,ha2016toward} who showed that multilingual models are somewhat capable of between untrained language pairs. Being able to translate between language pairs that were never trained on, these models give evidence for cross-lingual transfer. Zero-shot translation thus also becomes an important measure of model generalization. Unfortunately, the performance of such zero-shot translation is often not good enough to be useful as it is easily beaten by the simple pivoting approach. E.g., the multilingual model of \newcite{johnson2016google} scores 6 BLEU points lower on zero-shot Portuguese-to-Spanish translations.
    
    Since then there have been efforts to improve the quality of zero-shot translation. \newcite{DBLP:journals/corr/abs-1711-07893,ha2016toward} report a language bias problem in zero-shot translation wherein the multilingual NMT model often decodes to the wrong language. Strategies proposed to counter this include alternative ways of indicating the desired language, dictionary-based filtering at inference time, and balancing the dataset with additional monolingual data. While these techniques improved the quality of zero-shot translation, it was still behind pivoting by 4-5 BLEU points.
    
    There have also been several innovative proposals to promote cross-lingual transfer, the key to zero-shot translation, by modifying the model's architecture and selectively sharing parameters. \newcite{DBLP:conf/emnlp/FiratSAYC16} use separate encoders and decoders per language, but employ a common attention mechanism. In contrast, \newcite{blackwood2018multilingual} propose sharing all parameters but the attention mechanism. \newcite{platanios18emnlp} develop a contextual parameter generator that can be used to generate the encoder-decoder parameters for any source-target language pair. \newcite{lu2018neural} develop a shared ``interlingua layer'' at the interface of otherwise unshared, language-specific encoders and decoders. While making great progress, these efforts still end up behind pivoting by 2-10 BLEU points depending on the specific dataset and language pair. The above results suggest that parameter sharing alone is not sufficient for the system to learn language agnostic representations. 
    
    \paragraph{Language Invariance} as part of the the objective function may help solve this problem. Learning coordinated representations with the use of parallel data has been explored thoroughly in the context of multi-view and multi-modal learning \cite{wang2015deep,baltruvsaitis2018multimodal}. Without access to parallel data that can be used for direct alignment, a large mass of work minimizes domain discrepancy at the feature distribution level \citep{ben2007analysis,pan2011domain,ganin2016domain} to improve transfer. These techniques have been widely applied to learning cross-lingual representations \cite{mikolov2013distributed,hermann-blunsom:2014:P14-1}.
    
    The idea of aligning intermediate representations has also been explored in low-resource and unsupervised settings. \newcite{gu-EtAl:2018:N18-1} develop a way to align word embeddings and support cross-lingual transfer to languages with different scripts. They also apply a mixture-of-experts layer on top of the encoder to improve sentence level transfer, but this can be considered under the umbrella of parameter sharing techniques. Similar to our work, \newcite{artetxe2018unsupervised,DBLP:journals/corr/abs-1711-00043,DBLP:journals/corr/abs-1804-09057} explore applying adversarial losses on the encoder to ensure that the representations are language agnostic. However, more recent work on unsupervised NMT~\citep{DBLP:journals/corr/abs-1804-07755} has shown that the cycle consistency loss was the key ingredient in their systems. Such translation consistency losses have also been explored in \newcite{DBLP:journals/corr/abs-1805-04813,DBLP:journals/corr/abs-1805-10338,DBLP:journals/corr/XiaHQWYLM16}
    
    \paragraph{Zero-Resource NMT} are another class of methods to build translation systems for language pairs with no available training data. Unlike zero-shot translation systems, they are not immediately concerned with improving cross-lingual transfer. They instead address the problem by synthesizing a pseudo-parallel corpus that covers the missing parallel source-target data. This data is typically acquired by translating the English portion of available English-Source or English-Target parallel data to the third language \cite{DBLP:conf/emnlp/FiratSAYC16,DBLP:journals/corr/ChenLCL17}. With the help of this supervision, these approaches perform very well -- often beating pivoting. However, this style of zero-resource translation requires multiple phases to train teacher models, generate pseudo-parallel data (back-translation \cite{sennrich2015improving}), and then train a multilingual model on all possible language pairs. The added training complexity, along with the fact that it scales quadratically with the number of languages, makes these approaches less suitable for a truly multilingual setting.
    
\section{An Error Analysis of Zero-Shot Translation}
\label{error-analysis}
    For zero-shot translation to work, the intermediate representations of the multilingual model need to be language invariant. In this section we evaluate the degree to which a standard multilingual NMT system is able to achieve language invariant representations. We compare its translation quality to bilingual systems on both supervised and unsupervised (zero-shot) directions and develop an understanding of the pathologies that lead to low zero-shot quality.

\subsection{Experimental Setup}
\subsubsection{Data}
    Our experiments use the standard WMT14 en$\rightarrow$fr (39M) and en$\rightarrow$de (4.5M) training datasets that are used to benchmark state-of-the-art NMT systems \cite{DBLP:journals/corr/VaswaniSPUJGKP17,gehring2017convolutional,chen2018best}. We pre-process the data by applying the standard Moses pre-processing scripts.\footnote{We use {\tt normalize-punctuation.perl}, {\tt remove-non-printing-char.perl}, and {\tt tokenizer.perl}.} We swap the source and target to get parallel data for the fr$\rightarrow$en and de$\rightarrow$en directions. The resulting datasets are merged by oversampling the German portion to match the size of the French portion. This results in a total of 158M sentence pairs. The vocabulary is built by applying 32k BPE \citep{SennrichHB15} to obtain subwords. It is shared by both the encoder and the decoder. The target language \textless\textit{tl}\textgreater~tokens are also added to the vocabulary. For evaluation we use the 3-way parallel newstest-2012 (3003 sentence) as the dev set and newstest-2013 (3000 sentences) as the test set.\footnote{We could not use newstest-2014 since it is not 3-way parallel and would have made evaluating and analyzing results on de$\leftrightarrow$fr translation difficult.}
    
    \subsubsection{Model and Optimization}
    We run all our experiments with Transformers~\citep{DBLP:journals/corr/VaswaniSPUJGKP17}, using the TransformerBase configuration. Embeddings are initialized from a Gaussian distribution with scale $1/\sqrt{1024}$. We train our model with the transformer learning rate schedule using 4k warmup steps. Dropout is set to 0.1. We use synchronized training with 16 Tesla P100 GPUs and train the model until convergence, which takes around 500k steps. All models are implemented in Tensorflow-Lingvo \cite{shen2019lingvo}.
    
    The bilingual NMT models are trained as usual. Similar to \cite{johnson2016google}, our multilingual NMT model has the exact same architecture as the single direction models, using a single encoder and a single decoder for all language pairs. This setup maximally enforces the parameter sharing constraint that previous works rely on to promote cross-lingual transfer. Its simplicity also makes it favorable to analyze. The model is instructed on which language to translate a given input sentence into by feeding in a \textless\textit{tl}\textgreater~token, which is unique per target language, along with the source sentence.

\subsection{Baseline Result}

    \begin{table*}
    \centering
    \setlength\tabcolsep{3.5pt}
    \begin{tabular}{|l|c|c|c|c|c|c|c|c|} 
    \hline
    \multirow{2}{*}{\textbf{System}} & \multicolumn{2}{c|}{\textbf{Zero Shot}}                                                      & \multicolumn{2}{c|}{\textbf{Pivot}} & \multicolumn{4}{c|}{\textbf{Supervised}}                  \\ 
    \cline{2-9}
                            & \multicolumn{1}{c|}{$de\rightarrow fr$} & \multicolumn{1}{c|}{$fr\rightarrow de$} & \multicolumn{1}{c|}{$de\rightarrow fr$} & \multicolumn{1}{c|}{$fr\rightarrow de$} & \multicolumn{1}{c|}{$en\rightarrow fr$} & \multicolumn{1}{c|}{$en\rightarrow de$} & \multicolumn{1}{c|}{$fr\rightarrow en$} & \multicolumn{1}{c|}{$de\rightarrow en$}  \\ 
    \hline \hline
    single             & -                               & -                               & 27.59                                        & 20.71                                        & 34.76                                    & 24.31                                      & 33.61                             & 30.46                                     \\ 
    \hline \hline
    multi (drop=0.1)                 & 17.00                              & 11.84                               & \multicolumn{1}{c|}{26.25}               & \multicolumn{1}{c|}{20.18}               & 32.68                                    & 24.48                                    & 32.33                                    & 30.26                                     \\ 
    \hline
    multi (drop=0.3)                 & 21.57                               & 13.18                               & \multicolumn{1}{c|}{-}               & \multicolumn{1}{c|}{-}               & 29.64                                    & 21.98                                    & 29.55                                    & 27.52                                     \\ 
    \hline
    \end{tabular}
    \caption{WMT14 en-de-fr Zero-shot results with baseline and aligned models compared against pivoting.
    Pivoting through English is performed using the baseline multilingual model.}
    \label{table:base}
    \end{table*}

    We train 4 one-to-one translation models for $en\rightarrow de$, $en\rightarrow fr$, $de\rightarrow en$, and $fr\rightarrow en$, and one multilingual model for $en \leftrightarrow {de,fr}$ and report results in Table~\ref{table:base}. We see that the multilingual model performs well on the directions for which it received supervision: $en\leftrightarrow {de,fr}$. The 1-2 BLEU point regression as compared the one-to-one models is expected given that the multilingual model is trained to perform multiple tasks while using the same capacity.
    
    Pivoting results for $de\rightarrow fr$ and $fr \rightarrow de$ were obtained by first translating from German/French to English, and then translating the English to French/German. Once again the multilingual model performs well. The 1 BLEU drop as compared to the single model baseline arises from the relative difference in performance on supervised directions. Unlike single language pair models, the multilingual model is capable of zero-shot translation. Unfortunately, the quality is far below that of pivoting, making zero-shot translation unusable.

\subsection{Target Language is Entangled with Source Language}

    \begin{table}[h]
    \begin{center}
    \begin{tabular}{|l|c|c|c|}
    \hline
    ~ & en  & de  & fr \\
    \hline 
    $de \rightarrow fr$                  & 14\% & 25\% & 60\% \\
    \hline
    $fr \rightarrow de$                  & 12\% & 54\% & 34\% \\
    \hline
    \end{tabular}
    \caption{Percentage of sentences by language in reference translations and the sentences decoded using the baseline multilingual model (newstest2012)}
    \label{tab:langid}
    \end{center}
    \end{table}
    
    Inspecting the model's predictions, we find that a significant fraction of the examples were translated to the wrong language. They were either translated to English or simply copied as shown in Table~\ref{tab:langid}. This phenomenon has been reported before \cite{DBLP:journals/corr/abs-1711-07893,ha2016toward}. It is likely a consequence of the fact that at training time, German and French sentences were always translated into English. As a result, the model never learns to properly attribute the target language to the \textless\textit{tl}\textgreater~token, and simply changing the \textless\textit{tl}\textgreater~token at test time is not effective.
    
\subsection{Problems with Cross-lingual Generalization}
    \label{dropout-expts}
    \begin{table}[h]
    \begin{center}
    \begin{tabular}{|l|c|c|c|}
    \hline
    \multicolumn{1}{|l}{~} & \# examples & Pivot & Zero-Shot\\
    \hline
    $de \rightarrow fr$ & 1875/3003 & 19.71 & 19.22 \\       
    \hline
    $fr \rightarrow de$ & 1591/3003 & 24.33 & 21.63 \\
    \hline
    \end{tabular}
    \end{center}
    \caption{ BLEU on subset of examples predicted in the right language by zero-shot translation through the multilingual model (newstest2012)}
    \label{langid_bleu}
    \end{table}
    
    Given that a large portion of the errors are due to incorrect language, we try to estimate the improvement to zero-shot translation quality that could potentially be achieved by solving this issue. We discount these errors by re-evaluating the BLEU scores of zero-shot translation and pivoting on only those examples that the multilingual model already zero-shot translates to the right language. The results are shown in Table \ref{langid_bleu}. We find that although the vanilla zero-shot translation system is much stronger than expected at first glance, it still lags the pivoting by 0.5 BLEU points on French to German and by 2.7 BLEU points on German to French. This gap and the below analysis indicate a generalization problem in the model.

    One way to improve model generalization is by restricting the capacity of the model. With lower capacity, the model is expected be forced to learn cross-lingual representations which can be more broadly used across the different tasks of translating in many directions. This can be done by decreasing the number of parameters, for example through weight tying as previous multilingual approaches have done, or by simply increasing the regularization. We increase the dropout applied to the model from 0.1 to 0.3. We see that this results in higher zero-shot performance. However, this comes at high cost to the performance on supervised directions which end up being over-regularized.
 
\begin{figure*}[ht!]
    \centering
    \centerline{\includegraphics[width=\textwidth]{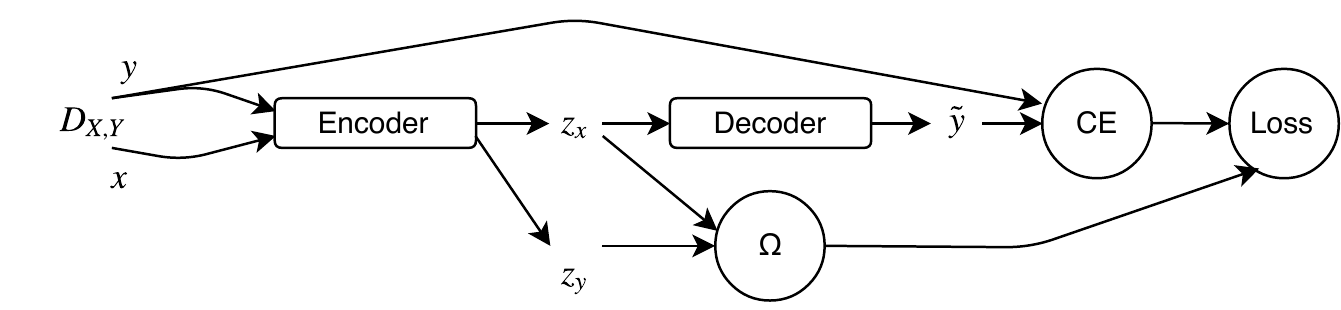}}
    \caption{The proposed multilingual NMT model along with alignment. $x$ and $y$ are a pair of translations sampled from available data, $D_{X,Y}$. One of $x$ or $y$ is always English. $z_x$ and $z_y$ are the encoder representations of the $x$ and $y$, respectively. $\tilde{y}$ is the decoder prediction. $CE$ is the standard cross-entropy loss associated with maximum likelihood training for NMT. $\Omega$ is the alignment loss. Both, $CE$ and $\Omega$, losses are minimized simultaneously.}
    \label{fig:algo}
\end{figure*} 
    
    Based on this, it seems that when a model is trained on just the end-to-end translation objective, there is no guarantee that it will discover language invariant representations; given enough capacity, it is possible for the model to partition its intrinsic dimensions and overfit to the supervised translation directions. Without any explicit incentive to learn invariant features, the intermediate encoder representations are specific to individual languages and this leads to poor zero-shot performance. While constraining model capacity can help alleviate this problem, it also impairs performance on supervised translation directions. We thus need to develop a more direct approach to push the model to learn transferable features.

\section{Aligning Latent Representations}
    
    To improve generalization to other languages, we apply techniques from domain adaptation to multilingual NMT. Multilingual NMT can be seen as a multi-task, multi-domain problem. Each source language forms a new domain, and each target language is a different task. We simplify this to a two domain problem by taking English to be the source domain, $D_{En}$, and grouping the non-English languages into the target domain, $D_{T}$. English is chosen as the source domain since it is the only domain for which we consistently have enough data for all the tasks/target languages. Minimizing the discrepancy between the feature distributions of the source and target domains will allow us to enable zero-shot translation \cite{ben2007analysis,mansour2009domain}. To this end we apply a regularizer while training the model that will force the model to make the representations of sentences in all non-English languages similar to their English counterparts - effectively making the model domain/source-language agnostic. In this way, English representations at the final layer of the encoder now form an implicit pivot in the latent space. The multilingual model is now trained on both, the cross-entropy translation loss and the new regularization loss. The loss function we then minimize is:
    
    \begin{equation}
    Loss = CE + \lambda \Omega
    \end{equation}

    \noindent where $CE$ is the cross-entropy translation loss, $\Omega$ is the alignment regularizer that will be defined below, and $\lambda$ is a hyper-parameter that controls the contribution of the alignment loss.
    
    Since we wish to make the representations source language invariant, we choose to apply the above regularization on top of the NMT encoder. This is because NMT models naturally decompose into an encoder and a decoder with a presumed separation of roles: The encoder encodes text in the source language into an intermediate latent representation, and the decoder generates the target language text conditioned on the encoder representation \cite{DBLP:journals/corr/ChoMGBSB14}.
    
    Below, we discuss two classes of regularizers that can be used. The first minimizes distribution level discrepancy between the source and target domain. The second uses the available parallel data to directly enforce a correspondence at the instance level .
    
\subsection{Aligning Distributions}
    We minimize the discrepancy between the feature distributions of the source and target domains by explicitly optimizing the following domain adversarial loss\citep{ganin2016domain}:

    \begin{equation}
    \begin{split}
    \Omega_{adv}&(\theta_{disc}) = \\
    & - \mathbb{E}_{\mathbf{x_{en}} \sim D_{En}}[-\log Disc(Enc(\mathbf{x_{en}}))] \\
    & + \mathbb{E}_{\mathbf{x_t} \sim D_T}[-\log (1-Disc(Enc(\mathbf{x_t})))]
    \end{split}
    \end{equation}
       
    \noindent where $Disc$ is the discriminator and is parametrized by $\theta_{disc}$. Note that, unlike \citet{artetxe2018unsupervised,DBLP:journals/corr/abs-1804-09057}, who also train their encoder adversarially to a language detecting discriminator, we are trying to align the distribution of encoder representations of all other languages to that of English and vice-versa. Our discriminator is just a binary predictor, independent of how many languages we are jointly training on.
    
    Architecturally, the discriminator is a feed-forward network with 3 hidden layers of dimension 2048 using the leaky ReLU($\alpha = 0.1$) non-linearity. It operates on the temporally max-pooled representation of the encoder output. We also experimented with a discriminator that made independent predictions for the encoder representation, $z_i$, at each time-step $i$ \cite{DBLP:journals/corr/abs-1711-00043}, but found the pooling based approach to work better for our purposes. More involved discriminators that consider the sequential nature of the encoder representations may be more effective, but we do not explore them in this work.

\subsection{Aligning Known Translation Pairs}
    The above adversarial domain adaptation strategy does not take full advantage of the fact that we have access to parallel data. Instead, it only enforces alignment between the source and the target domain at a distribution level. Here we attempt to make use of the available parallel data, and enforce an instance level correspondence between known translations, rather than just aligning the distributions in embedding space.
    
    Previous work on multi-modal and multi-view representation learning has shown that when given paired data, transferable representations can be much more easily learned by improving some measure of similarity between the alternative views \cite{baltruvsaitis2018multimodal}. In our case, the different views correspond to semantically equivalent sentences written in different languages. These are immediately available to us in our parallel training data. We now minimize:
    
    \begin{equation}
    \begin{split}
        & \Omega_{neg\_sim} = \\
        & -\mathbb{E}_{\mathbf{x_{t}},\mathbf{x_{en}} \sim D_{T,En}}[sim(Enc(\mathbf{x_t}), Enc(\mathbf{x_{en}}))]
    \end{split}
    \end{equation}
    
    \noindent where $D_{T,En}$ is the joint distribution of translation pairs. Note that $Enc(\mathbf{x_{t}})$ and $Enc(\mathbf{x_{en}})$ are actually a pair of sequences, and to compare them we would ideally have access to the word level correspondences between the two sentences. In the absence of this information, we make a bag-of-words assumption and align the pooled representation similar to \citet{gouws2015bilbowa,coulmance2016trans}. Empirically, we find that max pooling and minimizing the cosine distance between the representations of parallel sentences works well, but many other loss functions may yet be explored to obtain even better results.
    
\section{Experiments}
    We experiment with alignment on the same baseline multilingual setup as section \ref{error-analysis}. In addition to the model being trained end-to-end on the cross-entropy loss from translation, the encoder is also trained to minimize the alignment loss. To do this, we simultaneously encode both the source and the target sentence of all the translation pairs in a minibatch. While only the encoding of the source sentence is passed on to the decoder for translation, the encodings of both sentences are used to minimize the alignment loss.
    
    For cosine alignment, we simply minimize the cosine distance between the encodings of a given sentence pair. For adversarial adaptation, the encodings of all sentences in a batch are grouped into English and non-English encodings and fed to the discriminator. For each sentence encoding, the discriminator is trained to predict whether it came from the English group or the non-English group. On the other hand, the encoder is trained adversarially to the discriminator.
    
    $\lambda$ was tuned to 1.0 for both the adversarial and the cosine alignment loss. Simply fine-tuning a pre-trained multilingual model with SGD using a learning rate of 1e-4 works well, obviating the need to train from scratch. The models converge within a few thousand updates.

\subsection{Zero-Shot Now Matches Pivoting}
    \begin{table*}
    \centering
    \setlength\tabcolsep{4.1pt}
    \begin{tabular}{|l|c|c|c|c|c|c|} 
    \hline
    \multirow{2}{*}{\textbf{Multilingual System}} & \multicolumn{2}{c|}{\textbf{Zero Shot}}                                                      & \multicolumn{4}{c|}{\textbf{Supervised}}                  \\ 
    \cline{2-7}
                            & \multicolumn{1}{c|}{$de\rightarrow fr$} & \multicolumn{1}{c|}{$fr\rightarrow de$} &  \multicolumn{1}{c|}{$en\rightarrow fr$} & \multicolumn{1}{c|}{$en\rightarrow de$} & \multicolumn{1}{c|}{$fr\rightarrow en$} & \multicolumn{1}{c|}{$de\rightarrow en$}  \\ 
    \hline
    vanilla                 & 17.00                               & 11.84                                              & 32.68                                    & 24.48                                    & 32.33                                    & 30.26                                     \\ 
    \hline \hline
    adversarial             & 26.00                               & 20.39                                                                       & 32.92                                    & 24.5                                     & 32.39                                    & 30.21                                     \\ 
    \hline
    pool-cosine             & 25.85                               & 20.18                                                                       & 32.94                                    & 24.51                                    & 32.36                                    & 30.32                                     \\
    \hline
    \end{tabular}
    \caption{WMT14 en-de-fr Zero-shot results with baseline and aligned models compared against pivoting.
    Pivoting through English is performed using the baseline multilingual model.}
    \label{tab:main}
    \end{table*}
    
    We compare the zero-shot performance of the multilingual models against the pivoting with the same multilingual model. Pivoting was able to achieve BLEU scores of 26.25 on $de\rightarrow fr$ and 20.18 on $fr \rightarrow de$ as evaluated on newstest2013.  Our results in Table~\ref{tab:main} demonstrate that both our approaches to latent representation alignment result in large improvements in zero-shot translation quality for both directions, effectively closing the gap to the strong performance of pivoting. The alignment losses also effectively disentangle the representation of the source sentence from the target language ensuring prediction in the desired language.
    
    In contrast to naively constraining the model to encourage it to learn transferable representations as was explored in section \ref{dropout-expts}, the alignment losses are able to strike a much finer balance by taking a pin pointed approach to enforcing source language invariance. This is what allows us to push the model to generalize to the zero-shot language pairs without hurting the quality in the supervised directions.
    
\subsection{Quantifying the Improvement to Language Invariance}
    \begin{figure}[h]
    \centering
    \includegraphics[width=\textwidth,height=5cm,keepaspectratio]{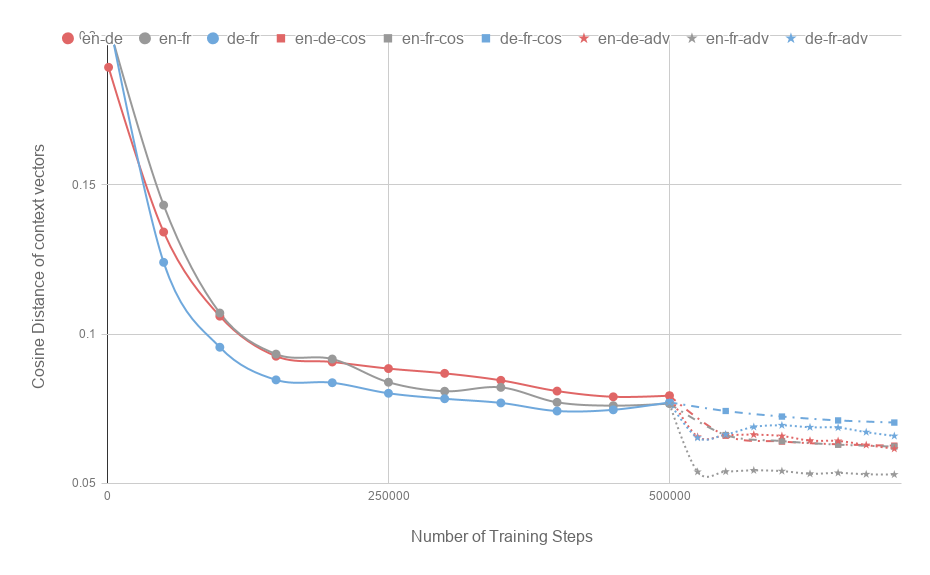}
    \caption{Average cosine distance between aligned context vectors for all combinations of English (en), German (de) and French (fr) as training progresses.}
    \label{fig:ctx-cos}
    \end{figure}
    
    We design a simple experiment to determine the degree to which representations learned while training a multilingual translation model are truly cross-lingual. Because sentences in different languages can have different lengths and word orders despite being translations of each other, it is not possible to directly compare encoder output representations. We instead go further downstream and compare the context vectors obtained while decoding from such a pair of sentences.
    
    In sequence-to-sequence models with attention, the attention mechanism is the only means by which the decoder can access the encoder representation. Thus, if we expect that for semantically equivalent source sentences, the decoder prediction should not change, then neither should the context vectors returned by the attention mechanism. Comparing context vectors obtained in such a manner will allow us to determine the extent to which their representations are functionally equivalent to the decoder.
    
    We sample a set of $100$ parallel en-de-fr sentences extracted from our dev set, newstest2012, for this analysis. For each sentence in each triple of aligned sentences, we obtain the sequence of pairs of context vectors while decoding to it from the other two sentences. We plot the mean cosine distances of these pairs for our baseline multilingual training run in Figure \ref{fig:ctx-cos}. We also show how these curves evolve when we fine-tune with the alignment losses. Our results indicate that the vanilla multilingual model learns to align encoder representations over the course of training. However, in the absence of an external incentive, alignment process arrests as training progresses. Incrementally training with the alignment losses results in a more language-agnostic representation which contributes to the improvements in zero-shot performance.
    
\subsection{Cosine vs Adversarial}
    The simple approach of just maximizing the representational similarity of known translation pairs is nearly indistinguishable from the quality of the more sophisticated adversarial training based approach. The adversarial regularizer suffers from three major problems: 1) it is sensitive to its initialization scheme and the choice of hyperparameters; 2) it has many moving parts coming from the architecture of the discriminator, the optimizer, and the non-linearity, all of which are non-trivial to control; and 3) it may also exhibit various failure modes including vanishing gradients and unstable oscillatory behaviour. In comparison, the cosine loss on translation pairs is simple, robust and effective with the only hyper-parameter being $\lambda$, which controls the weight of the alignment loss with respect to the translation loss.
    
\subsection{Zero-Shot with Adaptation vs Zero-Resource}
    We also evaluate against a zero-resource system. Here we synthesize $de \leftrightarrow fr$ parallel data by translating the $en$ portion of the available $en \leftrightarrow \{de,fr\}$ with previously trained one-to one models. We thus obtain 4.5M sentences with synthesized French and 39M sentences with synthesized German. These are reversed and concatenated to obtain 43.5M $de \leftrightarrow fr$ sentences. We then train two one-to-one NMT models for $de \rightarrow fr$ and $fr \rightarrow de$. These models obtained BLEU scores of 29.04 and 21.66 respectively on newstest2013. Note that these are one-to-one models and thus have the advantage of focussing on a single task as compared to a many-to-many multilingual model.
    
    While this approach achieves very good results it is hard to apply to a multilingual setting with many languages. It requires multiple phases: 1) Teacher models from English to each target language need to be trained, 2) Pseudo-parallel data for each language pair needs to be synthesized, 3) The multilingual model then needs to be jointly trained on data for all language pairs. The sequential nature of these phases and the quadratic scaling of this process make this approach unsuitable when we wish to support a large number of languages. In contrast, our approach does not require any additional pre-processing, additional training phases, or data generation. With the paired cosine alignment loss, the only hyper-parameter that we need to tune is $\lambda$.

\subsection{IWSLT17: Scaling to more languages}

    \begin{table}
    \centering
    \setlength\tabcolsep{3.8pt}

    \begin{tabular}{|l|c|c|c|} 
    \hline
    \multirow{2}{*}{\textbf{Group}}                   & \multicolumn{2}{c|}{\textbf{vanilla}}                             & \multicolumn{1}{c|}{\textbf{align(cosine)}}  \\ 
    \cline{2-4}
                                             & \multicolumn{1}{c|}{direct} & \multicolumn{1}{c|}{pivot} & \multicolumn{1}{c|}{direct}  \\ 
    \hline
    $en\leftrightarrow xx$ (8)  & 30.11                       & \multicolumn{1}{c|}{-}     & 29.95                        \\ 
    \hline
    $xx\leftrightarrow yy$ (12) & 16.73 (zs)                  & 17.76                      & 17.72 (zs)                   \\ 
    \hline
    All (20)                                 & 22.2                        & 22.81                      & 22.72                        \\
    \hline
    \end{tabular}
    \caption{Average BLEU scores for multilingual model on IWSLT-2017; Zero-Shot results are marked (zs).}
    \label{table:iwslt}
    \end{table}
    
    Here we demonstrate the scalability of our approach to multiple languages. We use the dataset from IWSLT-17 shared task which has transcripts of Ted talks in 5 languages: English~(en), Dutch~(nl), German~(de), Italian~(it), and Romanian~(ro). The original dataset is multi-way parallel with approximately 220 thousand sentences per language, but for the sake of our experiments we only use the to/from English directions for training. The dev and test sets are also multi-way parallel and comprise around 900 to 1100 sentences per language pair respectively. We again use the transformer base architecture but multiply learning rate by 2.0 and increase the number of warmup steps to 8k to make the learning rate schedule more conservative. Dropout is set to 0.2. We use the cosine loss with $\lambda$ set to 0.001, but higher values up to 0.1 are also equally effective. 
    
    On this dataset, the baseline model does not seem to have trouble with decoding to the correct language and performs well on zero-shot translation from the start. This may be a symptom of this dataset being multi-way parallel with the English sentences shared across all language pairs. However, it is still 1 BLEU point behind the quality of pivoting as shown in Table~\ref{table:iwslt}. By training with the auxillary cosine alignment loss, we are once again able to match the quality of bridging. 
    
\section{Conclusion}
   We started with an error-analysis of zero-shot translation in naively trained multilingual NMT and diagnosed why they do not automatically generalize to zero-shot directions. Viewing zero-shot NMT under the light of domain adaptation, we proposed auxillary losses to force the model to learn source language invariant representations that improve generalization. Through careful analyses we showed how these representations lead to better zero-shot performance while still maintaining performance on the supervised directions. We demonstrated the simplicity and effectiveness of our approach on two public benchmarks datasets: WMT English-French-German and the IWSLT 2017 shared task.

\section*{Acknowledgments}
We would like to thank the Google Brain and Google Translate teams for their useful inputs and discussions. We would also like to thank the entire Lingvo development team for their foundational contributions to this project. \\


\bibliography{acl2019}

\begin{thebibliography}{41}
\expandafter\ifx\csname natexlab\endcsname\relax\def\natexlab#1{#1}\fi

\bibitem[{Artetxe et~al.(2018)Artetxe, Labaka, Agirre, and
  Cho}]{artetxe2018unsupervised}
Mikel Artetxe, Gorka Labaka, Eneko Agirre, and Kyunghyun Cho. 2018.
\newblock \href {https://openreview.net/forum?id=Sy2ogebAW} {Unsupervised
  neural machine translation}.
\newblock In \emph{International Conference on Learning Representations}.

\bibitem[{Bahdanau et~al.(2014)Bahdanau, Cho, and
  Bengio}]{DBLP:journals/corr/BahdanauCB14}
Dzmitry Bahdanau, Kyunghyun Cho, and Yoshua Bengio. 2014.
\newblock \href {http://arxiv.org/abs/1409.0473} {Neural machine translation by
  jointly learning to align and translate}.
\newblock \emph{CoRR}, abs/1409.0473.

\bibitem[{Baltru{\v{s}}aitis et~al.(2018)Baltru{\v{s}}aitis, Ahuja, and
  Morency}]{baltruvsaitis2018multimodal}
Tadas Baltru{\v{s}}aitis, Chaitanya Ahuja, and Louis-Philippe Morency. 2018.
\newblock Multimodal machine learning: A survey and taxonomy.
\newblock \emph{IEEE Transactions on Pattern Analysis and Machine
  Intelligence}.

\bibitem[{Ben-David et~al.(2007)Ben-David, Blitzer, Crammer, and
  Pereira}]{ben2007analysis}
Shai Ben-David, John Blitzer, Koby Crammer, and Fernando Pereira. 2007.
\newblock Analysis of representations for domain adaptation.
\newblock In \emph{Advances in neural information processing systems}, pages
  137--144.

\bibitem[{Blackwood et~al.(2018)Blackwood, Ballesteros, and
  Ward}]{blackwood2018multilingual}
Graeme Blackwood, Miguel Ballesteros, and Todd Ward. 2018.
\newblock Multilingual neural machine translation with task-specific attention.
\newblock \emph{arXiv preprint arXiv:1806.03280}.

\bibitem[{Chen et~al.(2018)Chen, Firat, Bapna, Johnson, Macherey, Foster,
  Jones, Parmar, Schuster, Chen et~al.}]{chen2018best}
Mia~Xu Chen, Orhan Firat, Ankur Bapna, Melvin Johnson, Wolfgang Macherey,
  George Foster, Llion Jones, Niki Parmar, Mike Schuster, Zhifeng Chen, et~al.
  2018.
\newblock The best of both worlds: Combining recent advances in neural machine
  translation.
\newblock \emph{arXiv preprint arXiv:1804.09849}.

\bibitem[{Chen et~al.(2017)Chen, Liu, Cheng, and
  Li}]{DBLP:journals/corr/ChenLCL17}
Yun Chen, Yang Liu, Yong Cheng, and Victor O.~K. Li. 2017.
\newblock \href {http://arxiv.org/abs/1705.00753} {A teacher-student framework
  for zero-resource neural machine translation}.
\newblock \emph{CoRR}, abs/1705.00753.

\bibitem[{Cho et~al.(2014)Cho, van Merrienboer, G{\"{u}}l{\c{c}}ehre, Bougares,
  Schwenk, and Bengio}]{DBLP:journals/corr/ChoMGBSB14}
Kyunghyun Cho, Bart van Merrienboer, {\c{C}}aglar G{\"{u}}l{\c{c}}ehre, Fethi
  Bougares, Holger Schwenk, and Yoshua Bengio. 2014.
\newblock \href {http://arxiv.org/abs/1406.1078} {Learning phrase
  representations using {RNN} encoder-decoder for statistical machine
  translation}.
\newblock \emph{CoRR}, abs/1406.1078.

\bibitem[{Coulmance et~al.(2016)Coulmance, Marty, Wenzek, and
  Benhalloum}]{coulmance2016trans}
Jocelyn Coulmance, Jean-Marc Marty, Guillaume Wenzek, and Amine Benhalloum.
  2016.
\newblock Trans-gram, fast cross-lingual word-embeddings.
\newblock \emph{arXiv preprint arXiv:1601.02502}.

\bibitem[{Dong et~al.(2015)Dong, Wu, He, Yu, and Wang}]{dong2015multi}
Daxiang Dong, Hua Wu, Wei He, Dianhai Yu, and Haifeng Wang. 2015.
\newblock Multi-task learning for multiple language translation.
\newblock In \emph{Proceedings of the 53rd Annual Meeting of the Association
  for Computational Linguistics}, pages 1723--1732.

\bibitem[{Firat et~al.(2016{\natexlab{a}})Firat, Cho, and
  Bengio}]{DBLP:journals/corr/FiratCB16}
Orhan Firat, KyungHyun Cho, and Yoshua Bengio. 2016{\natexlab{a}}.
\newblock \href {http://arxiv.org/abs/1601.01073} {Multi-way, multilingual
  neural machine translation with a shared attention mechanism}.
\newblock \emph{CoRR}, abs/1601.01073.

\bibitem[{Firat et~al.(2016{\natexlab{b}})Firat, Sankaran, Al{-}Onaizan,
  Yarman{-}Vural, and Cho}]{DBLP:conf/emnlp/FiratSAYC16}
Orhan Firat, Baskaran Sankaran, Yaser Al{-}Onaizan, Fatos~T. Yarman{-}Vural,
  and Kyunghyun Cho. 2016{\natexlab{b}}.
\newblock \href {http://aclweb.org/anthology/D/D16/D16-1026.pdf} {Zero-resource
  translation with multi-lingual neural machine translation}.
\newblock In \emph{Proceedings of the 2016 Conference on Empirical Methods in
  Natural Language Processing, {EMNLP} 2016, Austin, Texas, USA, November 1-4,
  2016}, pages 268--277.

\bibitem[{Ganin et~al.(2016)Ganin, Ustinova, Ajakan, Germain, Larochelle,
  Laviolette, Marchand, and Lempitsky}]{ganin2016domain}
Yaroslav Ganin, Evgeniya Ustinova, Hana Ajakan, Pascal Germain, Hugo
  Larochelle, Fran{\c{c}}ois Laviolette, Mario Marchand, and Victor Lempitsky.
  2016.
\newblock Domain-adversarial training of neural networks.
\newblock \emph{The Journal of Machine Learning Research}, 17(1):2096--2030.

\bibitem[{Gehring et~al.(2017)Gehring, Auli, Grangier, Yarats, and
  Dauphin}]{gehring2017convolutional}
Jonas Gehring, Michael Auli, David Grangier, Denis Yarats, and Yann~N Dauphin.
  2017.
\newblock Convolutional sequence to sequence learning.
\newblock \emph{arXiv preprint arXiv:1705.03122}.

\bibitem[{Gouws et~al.(2015)Gouws, Bengio, and Corrado}]{gouws2015bilbowa}
Stephan Gouws, Yoshua Bengio, and Greg Corrado. 2015.
\newblock Bilbowa: Fast bilingual distributed representations without word
  alignments.
\newblock In \emph{International Conference on Machine Learning}, pages
  748--756.

\bibitem[{Gu et~al.(2018)Gu, Hassan, Devlin, and Li}]{gu-EtAl:2018:N18-1}
Jiatao Gu, Hany Hassan, Jacob Devlin, and Victor~O.K. Li. 2018.
\newblock \href {http://www.aclweb.org/anthology/N18-1032} {Universal neural
  machine translation for extremely low resource languages}.
\newblock In \emph{Proceedings of the 2018 Conference of the North American
  Chapter of the Association for Computational Linguistics: Human Language
  Technologies, Volume 1 (Long Papers)}, pages 344--354, New Orleans,
  Louisiana. Association for Computational Linguistics.

\bibitem[{Ha et~al.(2016)Ha, Niehues, and Waibel}]{ha2016toward}
Thanh-Le Ha, Jan Niehues, and Alexander Waibel. 2016.
\newblock Toward multilingual neural machine translation with universal encoder
  and decoder.
\newblock \emph{arXiv preprint arXiv:1611.04798}.

\bibitem[{Ha et~al.(2017{\natexlab{a}})Ha, Niehues, and
  Waibel}]{ha2017effective}
Thanh-Le Ha, Jan Niehues, and Alexander Waibel. 2017{\natexlab{a}}.
\newblock Effective strategies in zero-shot neural machine translation.
\newblock \emph{arXiv preprint arXiv:1711.07893}.

\bibitem[{Ha et~al.(2017{\natexlab{b}})Ha, Niehues, and
  Waibel}]{DBLP:journals/corr/abs-1711-07893}
Thanh{-}Le Ha, Jan Niehues, and Alexander~H. Waibel. 2017{\natexlab{b}}.
\newblock \href {http://arxiv.org/abs/1711.07893} {Effective strategies in
  zero-shot neural machine translation}.
\newblock \emph{CoRR}, abs/1711.07893.

\bibitem[{Hermann and Blunsom(2014)}]{hermann-blunsom:2014:P14-1}
Karl~Moritz Hermann and Phil Blunsom. 2014.
\newblock \href {http://www.aclweb.org/anthology/P14-1006} {Multilingual models
  for compositional distributed semantics}.
\newblock In \emph{ACL}, pages 58--68, Baltimore, Maryland.

\bibitem[{Johnson et~al.(2016)Johnson, Schuster, Le, Krikun, Wu, Chen, Thorat,
  Vi{\'e}gas, Wattenberg, Corrado et~al.}]{johnson2016google}
Melvin Johnson, Mike Schuster, Quoc~V Le, Maxim Krikun, Yonghui Wu, Zhifeng
  Chen, Nikhil Thorat, Fernanda Vi{\'e}gas, Martin Wattenberg, Greg Corrado,
  et~al. 2016.
\newblock Google's multilingual neural machine translation system: enabling
  zero-shot translation.
\newblock \emph{arXiv preprint arXiv:1611.04558}.

\bibitem[{Lample et~al.(2017)Lample, Denoyer, and
  Ranzato}]{DBLP:journals/corr/abs-1711-00043}
Guillaume Lample, Ludovic Denoyer, and Marc'Aurelio Ranzato. 2017.
\newblock \href {http://arxiv.org/abs/1711.00043} {Unsupervised machine
  translation using monolingual corpora only}.
\newblock \emph{CoRR}, abs/1711.00043.

\bibitem[{Lample et~al.(2018)Lample, Ott, Conneau, Denoyer, and
  Ranzato}]{DBLP:journals/corr/abs-1804-07755}
Guillaume Lample, Myle Ott, Alexis Conneau, Ludovic Denoyer, and Marc'Aurelio
  Ranzato. 2018.
\newblock \href {http://arxiv.org/abs/1804.07755} {Phrase-based {\&} neural
  unsupervised machine translation}.
\newblock \emph{CoRR}, abs/1804.07755.

\bibitem[{Lu et~al.(2018)Lu, Keung, Ladhak, Bhardwaj, Zhang, and
  Sun}]{lu2018neural}
Yichao Lu, Phillip Keung, Faisal Ladhak, Vikas Bhardwaj, Shaonan Zhang, and
  Jason Sun. 2018.
\newblock A neural interlingua for multilingual machine translation.
\newblock \emph{arXiv preprint arXiv:1804.08198}.

\bibitem[{Luong et~al.(2015)Luong, Le, Sutskever, Vinyals, and
  Kaiser}]{luong2015multi}
Minh-Thang Luong, Quoc~V Le, Ilya Sutskever, Oriol Vinyals, and Lukasz Kaiser.
  2015.
\newblock Multi-task sequence to sequence learning.
\newblock In \emph{International Conference on Learning Representations}.

\bibitem[{Mansour et~al.(2009)Mansour, Mohri, and
  Rostamizadeh}]{mansour2009domain}
Yishay Mansour, Mehryar Mohri, and Afshin Rostamizadeh. 2009.
\newblock Domain adaptation: Learning bounds and algorithms.
\newblock \emph{arXiv preprint arXiv:0902.3430}.

\bibitem[{Mikolov et~al.(2013)Mikolov, Sutskever, Chen, Corrado, and
  Dean}]{mikolov2013distributed}
Tomas Mikolov, Ilya Sutskever, Kai Chen, Greg~S Corrado, and Jeff Dean. 2013.
\newblock Distributed representations of words and phrases and their
  compositionality.
\newblock In \emph{Advances in neural information processing systems}, pages
  3111--3119.

\bibitem[{Pan et~al.(2011)Pan, Tsang, Kwok, and Yang}]{pan2011domain}
Sinno~Jialin Pan, Ivor~W Tsang, James~T Kwok, and Qiang Yang. 2011.
\newblock Domain adaptation via transfer component analysis.
\newblock \emph{IEEE Transactions on Neural Networks}, 22(2):199--210.

\bibitem[{Platanios et~al.(2018)Platanios, Sachan, Neubig, and
  Mitchell}]{platanios18emnlp}
Emmanouil~Antonios Platanios, Mrinmaya Sachan, Graham Neubig, and Tom Mitchell.
  2018.
\newblock \href {https://arxiv.org/abs/1808.08493} {Contextual parameter
  generation for universal neural machine translation}.
\newblock In \emph{Conference on Empirical Methods in Natural Language
  Processing (EMNLP)}, Brussels, Belgium.

\bibitem[{Ren et~al.(2018)Ren, Chen, Liu, Li, Zhou, and
  Ma}]{DBLP:journals/corr/abs-1805-04813}
Shuo Ren, Wenhu Chen, Shujie Liu, Mu~Li, Ming Zhou, and Shuai Ma. 2018.
\newblock \href {http://arxiv.org/abs/1805.04813} {Triangular architecture for
  rare language translation}.
\newblock \emph{CoRR}, abs/1805.04813.

\bibitem[{Salloum and Habash(2013)}]{salloum2013dialectal}
Wael Salloum and Nizar Habash. 2013.
\newblock Dialectal arabic to english machine translation: Pivoting through
  modern standard arabic.
\newblock In \emph{Proceedings of the 2013 Conference of the North American
  Chapter of the Association for Computational Linguistics: Human Language
  Technologies}, pages 348--358.

\bibitem[{Sennrich et~al.(2015)Sennrich, Haddow, and
  Birch}]{sennrich2015improving}
Rico Sennrich, Barry Haddow, and Alexandra Birch. 2015.
\newblock Improving neural machine translation models with monolingual data.
\newblock \emph{arXiv preprint arXiv:1511.06709}.

\bibitem[{Sennrich et~al.(2016)Sennrich, Haddow, and Birch}]{SennrichHB15}
Rico Sennrich, Barry Haddow, and Alexandra Birch. 2016.
\newblock Neural machine translation of rare words with subword units.
\newblock In \emph{Proceedings of the 54th Annual Meeting of the Association
  for Computational Linguistics}.

\bibitem[{Sestorain et~al.(2018)Sestorain, Ciaramita, Buck, and
  Hofmann}]{DBLP:journals/corr/abs-1805-10338}
Lierni Sestorain, Massimiliano Ciaramita, Christian Buck, and Thomas Hofmann.
  2018.
\newblock \href {http://arxiv.org/abs/1805.10338} {Zero-shot dual machine
  translation}.
\newblock \emph{CoRR}, abs/1805.10338.

\bibitem[{Shen et~al.(2019)Shen, Nguyen, Wu, Chen et~al.}]{shen2019lingvo}
Jonathan Shen, Patrick Nguyen, Yonghui Wu, Zhifeng Chen, et~al. 2019.
\newblock \href {http://arxiv.org/abs/1902.08295} {Lingvo: a modular and
  scalable framework for sequence-to-sequence modeling}.

\bibitem[{Sutskever et~al.(2014)Sutskever, Vinyals, and
  Le}]{DBLP:journals/corr/SutskeverVL14}
Ilya Sutskever, Oriol Vinyals, and Quoc~V. Le. 2014.
\newblock \href {http://arxiv.org/abs/1409.3215} {Sequence to sequence learning
  with neural networks}.
\newblock \emph{CoRR}, abs/1409.3215.

\bibitem[{Vaswani et~al.(2017)Vaswani, Shazeer, Parmar, Uszkoreit, Jones,
  Gomez, Kaiser, and Polosukhin}]{DBLP:journals/corr/VaswaniSPUJGKP17}
Ashish Vaswani, Noam Shazeer, Niki Parmar, Jakob Uszkoreit, Llion Jones,
  Aidan~N. Gomez, Lukasz Kaiser, and Illia Polosukhin. 2017.
\newblock \href {http://arxiv.org/abs/1706.03762} {Attention is all you need}.
\newblock \emph{CoRR}, abs/1706.03762.

\bibitem[{Wang et~al.(2015)Wang, Arora, Livescu, and Bilmes}]{wang2015deep}
Weiran Wang, Raman Arora, Karen Livescu, and Jeff Bilmes. 2015.
\newblock On deep multi-view representation learning.
\newblock In \emph{International Conference on Machine Learning}, pages
  1083--1092.

\bibitem[{Wu and Wang(2007)}]{Wu:2007:PLA:1452924.1452934}
Hua Wu and Haifeng Wang. 2007.
\newblock \href {https://doi.org/10.1007/s10590-008-9041-6} {Pivot language
  approach for phrase-based statistical machine translation}.
\newblock \emph{Machine Translation}, 21(3):165--181.

\bibitem[{Xia et~al.(2016)Xia, He, Qin, Wang, Yu, Liu, and
  Ma}]{DBLP:journals/corr/XiaHQWYLM16}
Yingce Xia, Di~He, Tao Qin, Liwei Wang, Nenghai Yu, Tie{-}Yan Liu, and
  Wei{-}Ying Ma. 2016.
\newblock \href {http://arxiv.org/abs/1611.00179} {Dual learning for machine
  translation}.
\newblock \emph{CoRR}, abs/1611.00179.

\bibitem[{Yang et~al.(2018)Yang, Chen, Wang, and
  Xu}]{DBLP:journals/corr/abs-1804-09057}
Zhen Yang, Wei Chen, Feng Wang, and Bo~Xu. 2018.
\newblock \href {http://arxiv.org/abs/1804.09057} {Unsupervised neural machine
  translation with weight sharing}.
\newblock \emph{CoRR}, abs/1804.09057.

\end{thebibliography}
\bibliographystyle{acl_natbib}

\end{document}